\pdfoutput=1
\documentclass[11pt]{article}

\usepackage[final]{acl}

\usepackage{times}
\usepackage{latexsym}

\usepackage[T1]{fontenc}
\usepackage[utf8]{inputenc}

\usepackage{microtype}

\usepackage{inconsolata}

\usepackage{graphicx}
\usepackage{amsmath}
\usepackage{booktabs}
\usepackage{multirow}
\usepackage{subcaption}
\usepackage{float}
\usepackage{placeins}
\usepackage{mathrsfs}
\usepackage[most]{tcolorbox}

\title{RJE: A Retrieval-Judgment-Exploration Framework for Efficient Knowledge Graph Question Answering with LLMs}

\author{
 \textbf{Can Lin\textsuperscript{1}\thanks{~Equal contribution.}},
 \textbf{Zhengwang Jiang\textsuperscript{1}\footnotemark[1]},
 \textbf{Ling Zheng\textsuperscript{1}},
 \textbf{Qi Zhao\textsuperscript{1}},
\\
 \textbf{Yuhang Zhang\textsuperscript{1,2}},
 \textbf{Qi Song\textsuperscript{1,3}\thanks{~Corresponding author.}},
 \textbf{Wangqiu Zhou\textsuperscript{4}}
\\
 \textsuperscript{1}University of Science and Technology of China
\\
 \textsuperscript{2}City University of Hong Kong
\\
 \textsuperscript{3}Deqing Alpha Innovation Institute, Huzhou, China
\\
 \textsuperscript{4}Hefei University of Technology
\\
  \small{
   \{can.lin, jzw02, lingzheng, zq2021, yhzhang\}@mail.ustc.edu.cn
 }
 \\
\small{
   qisong09@ustc.edu.cn, rafazwq@hfut.edu.cn
 }
}

\begin{document}
\maketitle
\begin{abstract}

Knowledge graph question answering (KGQA) aims to answer natural language questions using knowledge graphs.
Recent research leverages large language models (LLMs) to enhance KGQA reasoning, but faces limitations: retrieval-based methods are constrained by the quality of retrieved information, while agent-based methods rely heavily on proprietary LLMs.
To address these limitations, we propose Retrieval-Judgment-Exploration (RJE), a framework that retrieves refined reasoning paths, evaluates their sufficiency, and conditionally explores additional evidence. Moreover, RJE introduces specialized auxiliary modules enabling small-sized LLMs to perform effectively: Reasoning Path Ranking, Question Decomposition, and Retriever-assisted Exploration. 
Experiments show that our approach with proprietary LLMs (such as GPT-4o-mini) outperforms existing baselines while enabling small open-source LLMs (such as 3B and 8B parameters) to achieve competitive results without fine-tuning LLMs.
Additionally, RJE substantially reduces the number of LLM calls and token usage compared to agent-based methods, yielding significant efficiency improvements.\footnote{Code and data are available at: \url{https://github.com/Olgird/RJE}}

\end{abstract}

\section{Introduction}

Knowledge graph question answering (KGQA) aims to find answer entities from knowledge graphs (KGs) in response to natural language questions~\citep{jiangunikgqa}. With the development of open domain knowledge graphs, such as Freebase \citep{bollacker2008freebase} and Wikidata \citep{pellissier2016freebase}, KGQA has become an important research topic. 
Although multi-hop KGQA has been studied~\citep{zhang2022subgraph,ji2024retrieval,mavromatis2024gnn}, finding complex multi-hop reasoning edges to infer correct answers remains a challenge. 

Concurrently, large language models (LLMs) have demonstrated remarkable performance across various natural language processing (NLP) tasks \citep{huang2023towards,grattafiori2024llama}. This convergence has stimulated growing research into methodologies that leverage LLMs to enhance KGQA systems \citep{gao2023retrieval,zhang2025survey}. One straightforward approach is to fine-tune LLMs \citep{luoreasoning,jiang2024kg}, which incurs significant costs and risks catastrophic forgetting~\citep{luo2023empirical}. 
As illustrated in Figure~\ref{fig:framework}, an alternative class of methods does not require parameter modification of LLMs. They can be categorized as two types:

\begin{figure}[t]
  \includegraphics[width=\columnwidth]{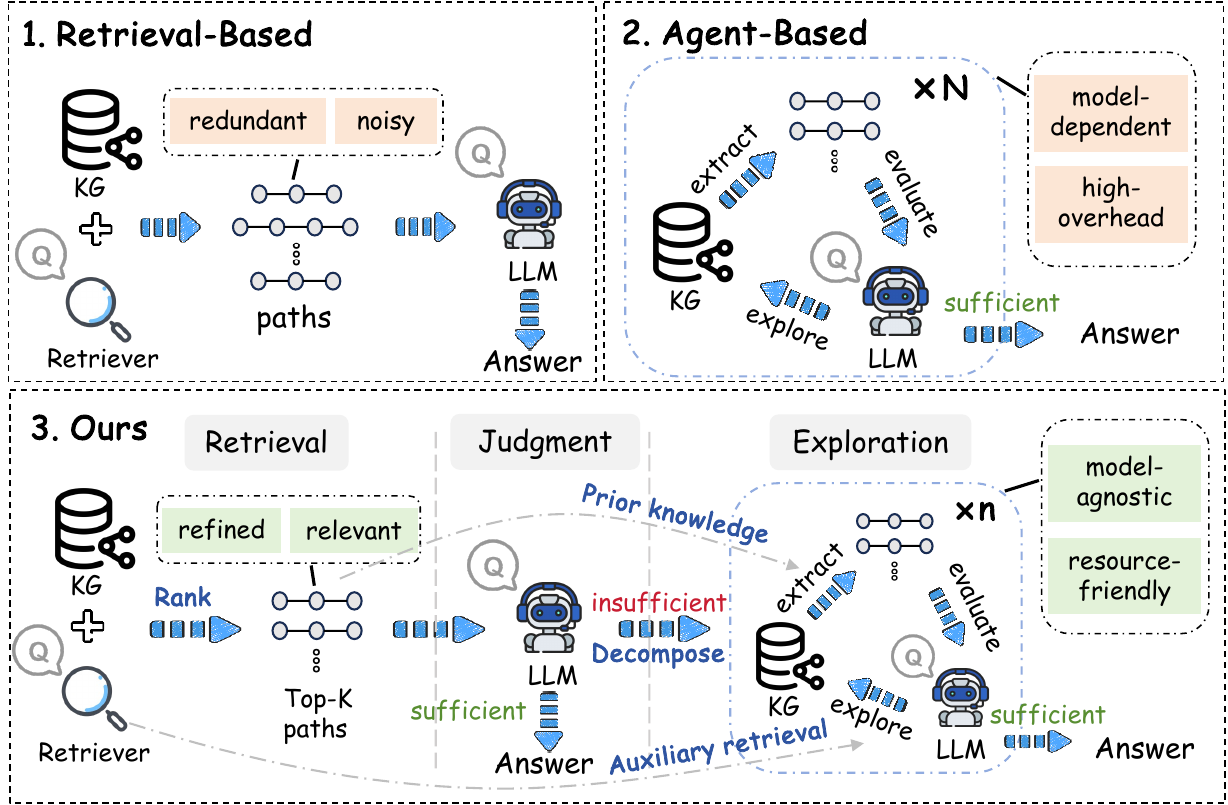}
  \caption{A comparison of three types of methods: retrieval-based methods, agent-based methods, and our proposed RJE that combines precise retrieval with conditional exploration.}% for more efficient question answering.}
  \label{fig:framework}
\end{figure}

1) \textbf{Retrieval-based methods} incorporate external retrieval mechanisms to extract relevant KG evidence, which is then integrated into textual prompts for LLMs to generate answers grounded in the retrieved knowledge \citep{he2024g,huang2024rd,li2025simple}.
However, \textbf{Constraints of retrieval information} limit the effectiveness of these methods. 
Retrieved information often struggles to simultaneously accommodate sufficiency and relevance.
Insufficient evidence prevents correct inference, while excessive information introduces noise that can derail the reasoning process~\citep{huang2024rd}.

2) \textbf{Agent-based methods} employ LLMs as agents that deliberately navigate the KG. Starting with question topic entities, LLMs systematically select candidate relations and entities, and then evaluate them in an iterative process until they reach an appropriate answer \citep{jiang2023structgpt,sunthink,chenplan}. These methods depend heavily on \textbf{proprietary LLMs}. They typically operate without initially provided external information, forcing LLMs to handle multi-step exploration independently. Additionally, exploration complexity increases substantially when reasoning over multiple topic entities and navigating extensive entities and relations in KGs. As a result, practical success has predominantly relied on proprietary LLMs like GPT-4, while smaller open-source alternatives demonstrate significantly lower performance \citep{li2024decoding}.

Above limitations led to a natural question: can we achieve \textit{more accurate reasoning with lower cost}? We answer this question by proposing a three-stage \textbf{R}etrieval-\textbf{J}udgment-\textbf{E}xploration (RJE) framework. As shown in Figure~\ref{fig:framework}, the RJE framework strategically synergizes KG retrieval with reasoning capabilities of LLMs by 
first retrieving refined reasoning paths from the KG, then employing LLMs to judge information sufficiency and finally executing targeted exploration as needed.

\textbf{Retrieval}: the retrieval stage of RJE prioritizes relevance over sufficiency of retrieved information. The retriever first extracts numerous reasoning paths from the KG based on relevance between the question and relations, where each reasoning path starts from a topic entity and consists of alternating relations and entities. To reduce noise and preserve the relevance, we introduce a lightweight \textit{Reasoning Path Ranking} module to further sort reasoning paths based on relevance scores, and only the top-$K$ reasoning paths are retained.

\textbf{Judgment}:
an LLM serves as a judge to evaluate whether the retained reasoning paths provide sufficient evidence to answer the question. If indeed sufficient, the LLM directly formulates a response; otherwise, an exploration phase will be executed to fetch the required information.

\textbf{Exploration}:
unlike prior agent-based methods \citep{sunthink,chenplan}, the exploration stage of RJE is dedicated to completing missing evidence and simplifying exploration.
When evidence is insufficient, the LLM acts as an exploration agent that first executes \textit{Question Decomposition} which breaks down a complex question into simpler sub-questions based on topic entities to focus on specific reasoning paths. 
RJE then utilizes retrieved reasoning paths as prior knowledge, allowing the LLM to initiate exploration from entities at knowledge gaps rather than topic entities, thereby reducing exploration steps. Subsequently, it iteratively conducts \textit{Retriever-assisted Exploration}, where the retriever pre-filters candidate relations to constrain the search space before the LLM conducts targeted relation and entity exploration to complete the missing evidence chain.

Our contributions can be summarized as follows:
\begin{itemize}
\item We introduce a novel framework called RJE that integrates precise retrieval, sufficiency judgment, and conditional exploration for KGQA. This framework simultaneously enhances reasoning capabilities and reduces computational demands.

\item We introduce three key auxiliary modules: Reasoning Path Ranking, Question Decomposition, and Retriever-assisted Exploration. These modules reduce the reasoning burden on LLMs, enabling small-sized LLMs to achieve performance comparable to prior work using proprietary models.

\item We conduct comprehensive experiments across standard KGQA benchmarks, demonstrating that RJE surpasses existing approaches in both accuracy and efficiency when using proprietary LLMs. Notably, with small open-source LLMs (3B and 8B parameters), RJE outperforms the previous state-of-the-art method PoG by 41.5\% and 27.9\% on the CWQ dataset under the same model sizes.

\end{itemize}

\section{Related Work}

The task of KGQA focuses on answering questions by leveraging information from KGs.
We categorize existing approaches as follows. 

\textbf{Semantic parsing methods} convert natural language questions into structured queries (e.g., SPARQL) or equivalent forms for execution \citep{cao2022program,hu2022logical, zhang2023fc, luo2024chatkbqa}. Despite their precision, these approaches often require costly logical form annotations and remain vulnerable to execution failures stemming from syntactic or semantic errors.

\textbf{Retrieval-based methods} typically comprise a retrieval module and a reasoning module. Early approaches \citep{zhang2022subgraph, jiangunikgqa,baek2023direct,li2023graph,jiang2023reasoninglm, ding2024enhancing} employed pre-trained models for retrieval paired with lightweight reasoning components. More recent work applies retrieval-augmented generation (RAG) with LLMs, utilizing retrieved subgraphs as contextual prompts \citep{baek2023knowledge, he2024g, zhao2024kg,mavromatis2024gnn, huang2024rd, li2025simple}.
However, these methods struggle to balance knowledge sufficiency and relevance, leading to either incomplete evidence for inference or noise that disrupts reasoning.

\textbf{Agent-based methods} extend beyond standard RAG implementations. Some studies explore fine-tuning LLMs to enhance their KG reasoning capabilities, such as RoG \citep{luoreasoning} and KG-Agent \citep{jiang2024kg}, though this strategy incurs significant computational costs and increases the risk of catastrophic forgetting \citep{luo2023empirical}. Alternatively, iterative prompting methods \citep{jiang2023structgpt,cheng2024call,markowitz2024tree,sunthink, chenplan,wang2025reasoning} leverage LLMs as agents to progressively retrieve relevant knowledge, with frameworks like ToG \citep{sunthink} and PoG \citep{chenplan} incorporating advanced strategies such as beam search, memory mechanisms, and reflection capabilities. However, these approaches often introduce additional system complexity, and small-sized LLMs continue to exhibit significant performance limitations.

\section{Preliminary}

\textbf{Knowledge Graph (KG)} is a structured semantic knowledge base, denoted as $\mathcal{G} = (E,R)$, where $E$ and $R$ represent the set of entities and relations respectively. A triple $\tau = (e, r, e')$ describes a fact with $e, e' \in E$ and $ r \in R$.

\textbf{Knowledge Graph Question Answering (KGQA)} aims to answer natural language questions using KGs. Formally, given a natural language question $q$ and a knowledge graph $\mathcal{G}$, the task of KGQA is to identify the corresponding answer entity subset ${A}_q \subseteq E$ that satisfies the question by leveraging the structural information in $\mathcal{G}$.

\textbf{Topic Entities} are the main entities mentioned in a question $q$ that serve as starting points for answer retrieval. Following prior work \citep{sunthink, chenplan, huang2024rd}, we assume that the set of topic entities $T$ has been identified from the question and successfully linked to nodes in KGs.

\textbf{Relation Path} is defined as a sequence of relations starting from a topic entity $e_t$. It is denoted as $p_r = (e_t, r_1, r_2, \ldots, r_{h})$, where $h$ represents the length of the relation path.

\textbf{Reasoning Path} refers to a complete path of alternating entities and relations, derived from a given relation path. It is formally represented as
$p_e = (e_t, r_1, e_1, r_2, \ldots, r_{h}, e_h)$, where each entity $e_{i-1}$ is connected to the next entity $e_{i}$ via relation $r_i$. Note that a single relation path may derive multiple reasoning paths based on the entity connections within the KG. For instance, $p'_e = (e_t, r_1, e'_1, r_2, \ldots, r_{h}, e'_h)$ constitutes an alternative reasoning path from the same relation path.

\section{Methodology}
\begin{figure*}[htbp]
  \centering
  \includegraphics[width=\textwidth]{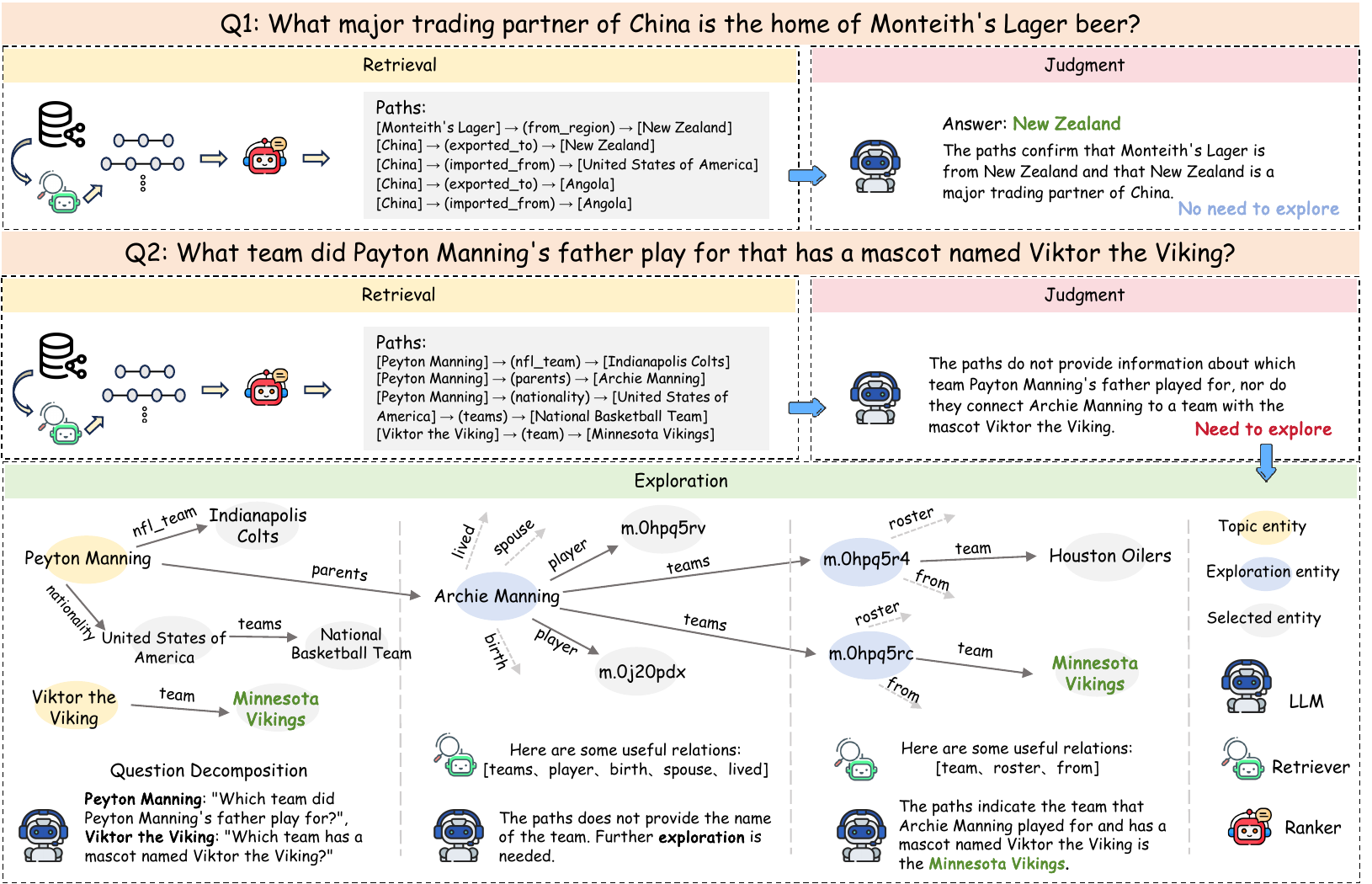}
  \caption{The framework overview of RJE, which flexibly adjusts its strategy based on the sufficiency of path evidence to minimize resource consumption while ensuring answer correctness.
  }
  \label{fig:main_p}
\end{figure*}

This section provides a detailed description of each stage in the RJE framework and its auxiliary modules. As illustrated in Figure~\ref{fig:main_p}, RJE consists of three stages: Retrieval, Judgment, and Exploration. Retrieval stage: the relation path retriever first identifies relevant relation paths from the KG, which are then refined by the reasoning path ranking module to obtain the top-$K$ reasoning paths.
Judgment stage: the top-$K$ reasoning paths are prompted to LLM, which evaluates whether the available evidence is sufficient to answer the question. If deemed sufficient, the LLM proceeds to generate an answer.
Exploration stage: when the current paths information is insufficient, further exploration is required. Firstly, the LLM performs question decomposition to guide further exploration. Then through an iterative process of exploration, the LLM gathers additional evidence until it determines sufficient information has been accumulated to generate the final answer.

\subsection{Retrieval Stage}

\subsubsection{Relation Path Retrieval}
In order to preliminarily extract information relevant to the question from the KG, we implement the relation path retrieval as the first step of retrieval stage.
Following the approach proposed by \citet{zhang2022subgraph} and \citet{huang2024rd}, we fine-tune a pre-trained language model as relation path retriever based on the question $q$, topic entities $T$, and answers $A$. This enables the model to capture the semantic similarity between the question and relevant relations. During retrieval, the model concatenates the question with the topic entity, employs beam search iteratively to assess semantic similarity between the question and neighboring candidate relations, and produces a set of relation paths $P_r = \{p_{r_0}, p_{r_1}, ..., p_{r_{\left| P_r \right|-1}}\}$, each with a corresponding relevance score, where $\left| P_r \right|$ is the number of selected relation paths.

\subsubsection{Reasoning Path Ranking} 
Depending on entity connections in the KG, a set of reasoning paths $P_e = \{p_{e_0}, p_{e_1}, ..., p_{e_{\left| P_e \right|-1}}\}$ is derived from $P_r$. To prevent overwhelming downstream LLMs with excessive reasoning paths, we introduce a reasoning path ranking module that employs a relatively small reasoning path ranker based on a pre-trained language model (PLM) to prioritize relevant reasoning paths.

For a specific reasoning path represented as $p_{e_i} = (e_t, r_1, e_1, r_2, e_2, ..., r_h, e_h)$, the ranker constructs an input sequence by concatenating the question and the path:
\begin{equation}
    \label{qi}
q_i = \{q \; [\text{SEP}] \; e_t \rightarrow r_1 \rightarrow e_1 \rightarrow \cdots \rightarrow e_h\}.
\end{equation}

Here, $[\text{SEP}]$ is the separator token used by the PLM, and the symbol ``$\rightarrow$'' indicates transition within the path. The sequence $q_i$ is fed into the PLM to obtain a relevance score:

\begin{equation}
    \label{qi}
s = \text{MLP}(E(q_i)),
\end{equation}
where $E(\cdot)$ denotes the embedding obtained from the PLM by extracting the $[\text{CLS}]$ token representation, and MLP is a multi-layer perceptron applied on top of the embedding. To ensure the relevance of extracted information and suppress irrelevant noise, we only select the top-$K$ reasoning paths based on relevance scores. As shown in Figure~\ref{fig:main_p}, for the question Q2 \textit{``What team did Payton Manning's father play for that has a mascot named Viktor the Viking?''}, only a few paths most relevant to the topic entities \textit{``Peyton Manning''} and \textit{``Viktor the Viking''} are selected. 

The ranker is trained with weak supervision, requiring only the question $q$ and the corresponding answer set $A$. For a reasoning path $p_{e_i}$, if there exists an answer $a \in A$ along that path, it is labeled as a positive sample. Otherwise, it is treated as a negative sample. The training objective adopts the margin ranking loss, defined as:
\begin{equation}
    \label{MRL}
\mathcal{L} = \max(0, s_{\text{neg}} - s_{\text{pos}} + \text{margin}),
\end{equation}
where $s_{\text{neg}}$ and $s_{\text{pos}}$ denote the scores for negative and positive samples, respectively, and $\text{margin}$ is a hyperparameter enforcing a minimum separation between them.

\subsection{Judgment Stage}
After extracting the top-$K$ reasoning paths, we employ an LLM-based judgment stage to assess information sufficiency. RJE feeds the paths into the LLM, which then evaluates whether these paths contain sufficient information to answer the question. If deemed sufficient, the LLM proceeds to generate answer. Otherwise, RJE identifies the need for further exploration. As shown in Figure~\ref{fig:main_p}, for the question Q1 \textit{``What major trading partner of China is the home of Monteith's Lager beer?''}, RJE successfully infers the correct answer by judging that the paths information is sufficient. In contrast, for the question Q2 \textit{``What team did Payton Manning's father play for that has a mascot named Viktor the Viking?''}, RJE identifies the insufficiency of the available information and proceeds with further exploration. These two examples demonstrate the flexibility and adaptability of the RJE framework. The prompt is provided in Appendix~\ref{prompt_Paths Evaluation}.

\subsection{Exploration Stage}
\subsubsection{Question Decomposition}
Given a question $q$ and its corresponding set of topic entities $T = \{ e_{t_1}, e_{t_2}, \dots , e_{t_n}\}$, the reasoning paths associated with different topic entities need to be jointly considered to infer the final answer. However, during the exploration phase, these paths are relatively independent of each other. To this end, RJE prompts the LLM to perform question decomposition, leveraging both the original question and each topic entity to generate a set of focused sub-questions. Specifically, for each topic entity $e_{t_i}$, a corresponding sub-question $q_{t_i}$ is generated, which we define as a topic question. This strategy encourages the LLM to concentrate on reasoning paths specific to each topic entity, thereby enhancing both the efficiency and accuracy of reasoning.  
The prompt is provided in Appendix~\ref{prompt_Question Decomposition}.

\subsubsection{Path Exploration}

To fetch the required information, RJE conducts further exploration over the KG.
Initially, based on the question decomposition, we obtain a set of topic questions $Q = \{ q_{t_1}, q_{t_2}, \dots, q_{t_n} \}$, where $n$ is the number of topic entities. The reasoning path ranker extracts the top-$K$ reasoning paths, which are categorized as $\mathcal{P}^0 = \{ P^0_{t_1}, P^0_{t_2}, \dots, P^0_{t_n} \}$. Here, $P^0_{t_i} = \{ p^0_{t_i,1}, p^0_{t_i,2}, \dots \}$ represents the set of paths originating from the topic entity $e_{t_i}$. 
We initialize the entity set $E^0$ as the collection of all entities that appear in the paths of $\mathcal{P}^0$. 
In the $D$-th round of exploration, assuming that the entity set $E^{D-1}$ and the paths $\mathcal{P}^{D-1}$ have already been obtained from the previous round, we perform further path exploration based on $E^{D-1}$ and $\mathcal{P}^{D-1}$ to acquire additional useful information from the KG.

\paragraph{Exploration Entities Selection.}
For different topic entities, the number of hops required to reach the answer entity can vary significantly. As shown in Figure~\ref{fig:main_p}, theoretically, for question Q2, the answer entity \textit{``Minnesota
Vikings''} can be reached in one hop from the topic entity \textit{``Viktor the Viking''}, whereas it takes three hops from \textit{``Peyton Manning''}. To ensure more efficient exploration, we prompt the LLM using the paths of the previous round $\mathcal{P}^{D-1}$, the original question $q$, the set of topic questions $Q$, and the set of topic entities $T$ to select a subset of entities $E^D_f$. For Q2, RJE selects the entity \textit{``Archie Manning''} in the first round and the entities \textit{``m.0hpq5r4''} and \textit{``m.0hpq5rc''} in the second round. 
This approach, combined with the path information, identifies which paths require further exploration to answer their corresponding topic questions, thereby avoiding redundant exploration on information-sufficient paths. 
The prompt is provided in Appendix~\ref{prompt_Exploration Entities Selection}.

\paragraph{Retriever-assisted Relation Exploration.}
Relation exploration aims to identify all relations relevant to answering the question.
Let $E^D_f = \{e^D_{f,1}, e^D_{f,2}, \dots, e^D_{f,m}\}$ denote the set of exploration entities selected in the $D$-th round. Each entity $e^D_{f,i}$ corresponds to a topic question $q_{t_c}$. We use pre-defined SPARQL queries to retrieve all relations connected to $e^D_{f,i}$, obtaining a candidate relation set $R^D_{f,i}$. 
For more reliable LLM reasoning, we employ the relation path retriever to filter the top-$N$ most relevant relations from $R^D_{f,i}$, resulting in a refined set $R^D_{s,i}$.
We then prompt the LLM with the topic question $q_{t_c}$, the entity $e^D_{f,i}$, and the relation set $R^D_{s,i}$ to select a subset of the most useful relations, denoted as $R^D_i$.
By performing this procedure for all entities in $E^D_f$, we derive the final set of explored relations in the $D$-th round $R^D = \{R^D_1, R^D_2, \dots, R^D_m\}$. The prompt and SPARQL queries are provided in Appendix~\ref{prompt_Relation Exploration} and Appendix~\ref{Relation_SPARQL}, respectively.

\paragraph{Entity Exploration.}
Entity exploration aims to further infer useful entities to help answer the question.
Given the entity set $E^D_f$ and the corresponding relation set $R^D$ obtained in the $D$-th round, we execute pre-defined SPARQL queries to retrieve the tail entities connected to each entity-relation pair. Following the approach of \citet{chenplan}, we apply a lightweight, train-free BERT to remove semantically irrelevant entities, resulting in a filtered entity set $E^D_s = \{ E^D_{s,1}, E^D_{s,2}, \dots, E^D_{s,k} \}$. 
Each $E^D_{s,i}$ is a set of tail entities derived from a specific pair $\langle e^D_{f,j}, r \rangle$, where $e^D_{f,j} \in E^D_f$ and $r \in R^D_j$. Let $q_{t_c}$ denote the topic question associated with $e^D_{f,j}$.
The LLM then performs entity selection based on the topic question $q_{t_c}$, the entity $e^D_{f,j}$, the relation $r$, and the candidate tail entity set $E^D_{s,i}$ to select a minimal subset of entities most valuable for answering the question, denoted as $E^D_i$.
After performing entity selection for all relevant entities and relations, we obtain the final entity set $E^D = \{E^D_1, E^D_2, \dots, E^D_k\}$. The prompt and SPARQL queries are provided in Appendix~\ref{prompt_Entity Exploration} and Appendix~\ref{Entity_SPARQL}, respectively.

\subsubsection{Answer Generation}

After performing relation and entity exploration in the $D$-th round, we use the relation set $R^D$ and the entity set $E^D$ to update and extend the paths from the previous round $\mathcal{P}^{D-1}$ to obtain $\mathcal{P}^D = \{P^D_{t_1}, P^D_{t_2}, \dots, P^D_{t_n} \}$. 
Given the original question $q$, the set of topic questions $Q = \{q_{t_1}, q_{t_2}, \dots, q_{t_n} \}$, and the updated paths $\mathcal{P}^D$, we prompt the LLM to reason over each path $P^D_{t_i}$ to answer the corresponding topic question $q_{t_i}$. Then, the LLM integrates the answers to the topic questions to infer the answer to the original question $q$.
During this process, if the LLM determines that the current paths provide sufficient information to answer the question, the iteration terminates and the final answer is generated. Otherwise, the LLM proceeds to perform a new round of path exploration.
To avoid endless exploration, we define a maximum number of exploration rounds $D_{\max}$. If the LLM is still unable to generate an answer after reaching $D_{\max}$, it will produce an answer based on the accumulated paths and its internal knowledge. The prompt is provided in Appendix~\ref{prompt_Answer Generation}.

\section{Experiments}
In this section, we present our experimental design, empirical results, and comprehensive analyses.
Our experiments address the following research questions (RQs):
\textbf{RQ1}:\ Does RJE achieve superior performance compared to state-of-the-art approaches on KGQA tasks?
\textbf{RQ2}:\ Can small-sized, open-source LLMs deliver competitive results within the RJE framework?
\textbf{RQ3}:\ Are the core stages and auxiliary modules of RJE effective in contributing to overall system performance?
\textbf{RQ4}:\ Can RJE reduce computational overhead and improve efficiency during reasoning?

\subsection{Datasets \& Evaluation Metrics}
To evaluate our proposed KGQA approach, we conduct extensive experiments on two widely used benchmark datasets that rely on the external knowledge graph Freebase \citep{bollacker2008freebase}: WebQuestionsSP (WebQSP) \citep{yih2016value} and Complex WebQuestions (CWQ) \citep{talmor2018web}. Detailed statistics of the datasets are provided in Appendix~\ref{sec:Datasets}.
Following prior research \citep{sunthink,chenplan,li2025simple}, we adopt Hits@1 (exact match accuracy) as our primary evaluation metric.

\subsection{Selected Baselines}
For a comprehensive comparison with various methods, we selected five categories of baselines as follows:
(1) \textit{LLM-only methods}: standard prompting (IO prompt) \citep{brown2020language}, Chain-of-Thought prompting (CoT) \citep{wei2022chain}, and Self-Consistency (SC) \citep{wangself}. (2) \textit{Retrieval-Reasoning Methods}: SR \citep{zhang2022subgraph}, UniKGQA \citep{jiangunikgqa}, ReasoningLM \citep{jiang2023reasoninglm} and EPR \citep{ding2024enhancing}. (3) \textit{Retrieval-Augmented Generation Methods}: RD-P \citep{huang2024rd}, KG-CoT \citep{zhao2024kg} and SubgraphRAG \citep{li2025simple}. (4) \textit{Fine-tuned Agent-based Methods}: RoG \citep{luoreasoning}, KG-Agent \citep{jiang2024kg} and GCR \citep{luo2024graph}. (5) \textit{Prompting Agent-based Methods}: StructGPT \citep{jiang2023structgpt}, ToG \citep{sunthink}, Interactive-KBQA \citep{xiong2024interactive}, ReKnoS \citep{wang2025reasoning} and PoG \citep{chenplan}. The description of the baselines can be found in Appendix~\ref{sec:appendix_Baselines}.

\subsection{Implementation Details}
We employ RoBERTa-base \citep{liu2019roberta} as our backbone PLM, consistent with prior work. For the reasoning path ranker training, we use a learning rate of 2e-5 with a margin of 1.0 for WebQSP, and 1e-5 with a margin of 0.8 for CWQ. Our framework supports integration with various LLMs, we evaluate with both proprietary models (ChatGPT, GPT-4o-mini, DeepSeek-V3) and open-source models (Llama3.2-3B, Llama3.1-8B, Qwen2.5-14B). Across all experiments, we set the number of reasoning paths to 10 ($K=10$), the number of filtered relations to 30 ($N=30$), LLM temperature to 0.3, and maximum exploration rounds $D_{\max}$ to 2 for WebQSP and 4 for CWQ. Proprietary LLMs were accessed through their official APIs\footnote{\url{https://platform.openai.com/docs/overview}}\footnote{\url{https://api-docs.deepseek.com}}, while open-source models were deployed on 4 NVIDIA A800-80G GPUs.

\subsection{Main Results (RQ1 \& RQ2)}

\begin{table}[t]
  \centering
  % \small
  \resizebox{\columnwidth}{!}{
  \begin{tabular}{lcc}
    \hline
    \textbf{Method} & \textbf{CWQ} & \textbf{WebQSP} \\
    \hline
    
    \multicolumn{3}{c}{\textit{LLM-Only methods}} \\
    \hline
    IO Prompt \citep{brown2020language}    & 37.6 & 63.3 \\
    CoT \citep{wei2022chain}       & 38.8 & 62.2 \\
     SC \citep{wangself}         & 45.4 & 61.1 \\
    \hline
    
    \multicolumn{3}{c}{\textit{Retrieval-Reasoning Methods}} \\
    \hline
    SR \citep{zhang2022subgraph}& 50.2 & 68.9 \\
    UniKGQA \citep{jiangunikgqa}    & 51.2 & 77.2 \\
    ReasoningLM \citep{jiang2023reasoninglm}&69.0 &78.5 \\
    EPR \citep{ding2024enhancing} & 60.6 & 71.2\\
    \hline
    
    \multicolumn{3}{c}{\textit{Retrieval-Augmented Generation Methods}} \\
    \hline
     KG-CoT w/GPT-4 \citep{zhao2024kg}    & 62.3 & 84.9 \\
     RD-P w/ChatGPT \citep{huang2024rd}    & 63.5 & 85.3 \\
     SubgraphRAG w/GPT4o \citep{li2025simple}     & 67.5 & 90.9 \\

    \hline
    
    \multicolumn{3}{c}{\textit{Fine-tuned Agent-based Methods}} \\
    \hline
     RoG \citep{luoreasoning}        & 62.6 & 85.7 \\
    % & Retrieval and Reasoning\cite{ji2024retrieval}        & 68.7 & 91.5 \\
     KG-Agent \citep{jiang2024kg}       & 72.2 & 83.3 \\
     GCR \citep{luo2024graph}       & 75.8 & 92.2 \\
    \hline
    
    \multicolumn{3}{c}{\textit{Prompting Agent-based Methods}} \\
    \hline
    StructGPT w/ChatGPT \citep{jiang2023structgpt}   & 54.3 & 72.6 \\
    Interactive-KBQA w/GPT-4 \citep{xiong2024interactive}    & 59.2 & 72.5 \\
     ToG w/ChatGPT \citep{sunthink}        & 58.9 & 76.2 \\
     ToG w/GPT-4 \citep{sunthink}        & 69.5 & 82.6 \\
     ReKnoS w/GPT-4o-mini \citep{wang2025reasoning}    & 66.8 & 83.8 \\
     PoG w/ChatGPT \citep{chenplan}        & 63.2 & 82.0 \\
     PoG w/GPT-4 \citep{chenplan}  & 75.0 & 87.3    \\
    \hline
    
    \multicolumn{3}{c}{\textit{Ours}} \\
    \hline
    RJE w/Llama3.2-3B   & 62.9 & 82.6\\
    RJE w/Llama3.1-8B   & 71.5 & 89.2\\
    RJE w/Qwen2.5-14B   & 71.2 & 90.3\\
    RJE w/ChatGPT        & 72.6 & 91.2\\
    RJE w/GPT-4o-mini    & 77.1 & \textbf{92.5}\\
    RJE w/DeepSeek-V3    & \textbf{78.2} & 92.4 \\
    \hline
  \end{tabular}
  }
  \caption{Results of different methods and models on two datasets. The best results are highlighted in bold.}
  \label{tab:main_result}
\end{table}

\begin{table}[t]
\centering
\small
\begin{tabular}{llcc}
\toprule
\textbf{Model} & \textbf{Method} & \textbf{CWQ} & \textbf{WebQSP} \\
\midrule
\multirow{3}{*}{Llama3.2-3B} 
& ToG     & 17.6 &  40.2\\
& PoG  & 21.4 & 49.3\\
& \textbf{RJE}  & \textbf{62.9} & \textbf{82.6} \\
\midrule
\multirow{3}{*}{Llama3.1-8B} 
& ToG     & 35.5 & 66.8 \\
& PoG  & 43.6 & 75.7 \\
& \textbf{RJE}  & \textbf{71.5} & \textbf{89.2} \\
\midrule
\multirow{3}{*}{Qwen2.5-14B} 
& ToG     & 39.0 & 75.3 \\
& PoG  & 54.3 & 79.6 \\
& \textbf{RJE}  & \textbf{71.2} & \textbf{90.3} \\
\midrule
\multirow{3}{*}{GPT-4o-mini} 
& ToG     & 65.4 & 80.7 \\
& PoG  & 67.2 & 82.4 \\
& \textbf{RJE}  & \textbf{77.1} & \textbf{92.5} \\
\bottomrule
\end{tabular}
\caption{Results of ToG, PoG and RJE on various backbone models on two datasets.}
\label{tab:various_model_result}
\end{table}

We compare RJE with various state-of-the-art baseline methods to evaluate its effectiveness in KGQA. 
As shown in Table~\ref{tab:main_result}, RJE consistently delivers performance gains across different LLMs and datasets. 
When GPT-4o-mini and DeepSeek-V3 are used as backbone LLMs, RJE outperforms all baseline methods, including LLM-Only methods, Retrieval-Reasoning Methods, Retrieval-Augmented Generation Methods, Fine-tuned Agent-based Methods, and Prompting Agent-based Methods. This highlights the superiority of our framework. It is noteworthy that small-sized LLMs also demonstrate competitive performance within our framework. For instance, Llama3.2-3B within RJE achieves performance comparable to ChatGPT used in PoG, the performance gap between the two is merely 0.3\% on CWQ. 
Meanwhile, RJE with Llama3.1-8B performs slightly below PoG with GPT-4 on CWQ but slightly surpasses it on WebQSP. 
Additionally, we present comprehensive Macro-F1 results in Appendix~\ref{sec:Analysis of Macro-F1 Metric}.

As shown in Table~\ref{tab:various_model_result}, RJE provides significant performance enhancement for small-sized LLMs.
Specifically, the smaller the LLM, the greater the performance boost from RJE. With Llama3.1-8B, RJE achieves a 27.9\% improvement over PoG on CWQ and 13.5\% on WebQSP. Even more notably, using Llama3.2-3B, RJE achieves a remarkable improvement of 41.5\% over PoG on CWQ (62.9\% vs. 21.4\%) and 33.3\% on WebQSP (82.6\% vs. 49.3\%).
Although the improvements are less dramatic with larger and more advanced LLMs, such as Qwen2.5-14B and GPT-4o-mini, RJE still consistently outperforms PoG by at least 9.9\% on both datasets.
These results suggest that RJE is particularly valuable for enhancing the capabilities of smaller LLMs, effectively narrowing the performance gap between smaller and larger models in KGQA tasks.
Additionally, Results Analysis and Case Study can be found in Appendix~\ref{sec:Results Analysis} and Appendix~\ref{sec:Case Study}, respectively.

\subsection{Ablation Study (RQ3)}

\begin{table}[t]
\centering
\small
\begin{tabular}{lcc}
\toprule
\textbf{Method} & \textbf{CWQ} & \textbf{WebQSP} \\
\midrule
RJE & \textbf{71.5} & \textbf{89.2} \\
w/o Exploration & 41.1 & 73.7 \\
w/o Retrieval & 60.7 & 83.9 \\
w/o Ranking & 66.9 & 88.0 \\
w/o Decomposition & 68.8 & 86.0 \\
w/o Assistance & 70.2 & 87.3 \\
\bottomrule
\end{tabular}
\caption{Ablation experiment results using Llama3.1-8B of removing each stage and each module, respectively.}
\label{table: ablation}
\end{table}

We conduct comprehensive ablation experiments to evaluate each component of our RJE framework, using Llama3.1-8B as the backbone LLM across both CWQ and WebQSP datasets. Table~\ref{table: ablation} presents the results of systematically removing key stages and modules: \textbf{w/o Exploration} (removing the exploration stage, using only the retrieval and judgment stages), \textbf{w/o Retrieval} (removing the retrieval stage, using only the exploration stage), \textbf{w/o Ranking} (removing the reasoning path ranking module), \textbf{w/o Decomposition} (removing the question decomposition module), and \textbf{w/o Assistance} (removing the retriever-assisted exploration module). 

As shown in Table~\ref{table: ablation}, the results demonstrate that each component contributes positively to the overall performance. 
Specifically, on CWQ, w/o Retrieval results in a 10.8\% drop in performance, while w/o Exploration leads to a more substantial 30.4\% drop. On WebQSP, performance decreases by 5.3\% and 15.5\% when w/o Retrieval and w/o Exploration, respectively. 
These findings highlight the critical importance of the synergistic collaboration between retriever and LLMs within our RJE framework.
The auxiliary modules in our framework contribute measurable performance improvements as well. Specifically, on WebQSP, the reasoning path ranking, question decomposition, and retriever-assisted exploration modules improve performance by 1.2\%, 3.2\%, and 1.5\%, respectively. On the more challenging CWQ dataset, the corresponding improvements are 4.6\%, 2.7\%, and 1.9\%. These results demonstrate that the auxiliary modules effectively alleviate the reasoning burden on small-sized LLMs in KGQA tasks, contributing to enhanced overall system performance.

Additional ablation studies on reasoning path ranking and the impact of the number of reasoning paths, filtered relations, and exploration rounds are provided in Appendix~\ref{sec:The Performance of Reasoning Path Ranker} and Appendix~\ref{sec:Addition Ablation study}.

\subsection{Efficiency Study (RQ4)}

\begin{table}[t]
\centering
\resizebox{\columnwidth}{!}{  
\begin{tabular}{cccccc}
\toprule
\textbf{Dataset} & \textbf{Method} & \textbf{LLM Call} & \textbf{Input Token} & \textbf{Output Token} & \textbf{Time (s)}\\
\midrule
\multirow{3}{*}{CWQ} & ToG & 22.6 & 8,182.9 & 1,486.4 & 96.5\\
 & PoG & 13.3 & 7,803.0 & 353.2 & 23.3\\

  & \textbf{RJE} & \textbf{7.9} & \textbf{5,769.1} & \textbf{247.2} & \textbf{16.3}\\
 
\midrule
\multirow{3}{*}{WebQSP} & ToG & 15.9 & 6,031.2 & 987.7 & 63.1 \\
 & PoG &  9.0 & 5,234.8 & 282.9 & 16.8 \\

 & \textbf{RJE} & \textbf{4.1} & \textbf{2763.3} & \textbf{148.7} & \textbf{10.5}\\
\bottomrule
\end{tabular}
} 
\caption{Efficiency comparison between our proposed RJE and baseline methods ToG and PoG.}
\label{tab:efficiency}
\end{table}

We evaluate the computational efficiency of RJE against two leading Prompting LLM approaches: ToG and PoG. Table~\ref{tab:efficiency} summarizes the efficiency metrics across the CWQ and WebQSP datasets, including the average number of LLM calls, token usage, and time consumption per question. As shown in Table~\ref{tab:efficiency}, RJE demonstrates superior efficiency in LLM calls, token utilization, and time consumption. 
On CWQ, RJE reduces LLM calls by 65.0\% compared to ToG and 40.6\% compared to PoG. This efficiency gain is even more pronounced on WebQSP, where RJE achieves reductions of 74.2\% and 54.4\% in LLM calls compared to ToG and PoG, respectively. 
As for token consumption, RJE reduces token usage by approximately 30\% compared to PoG on CWQ, and by about 50\% on WebQSP.
Additionally, RJE exhibits significant time efficiency improvements, performing at least 5.9 times faster than ToG and at least 1.4 times faster than PoG across both datasets.

The efficiency gains of RJE stem from two key design aspects: first, its judgment stage directly answers the question when the paths provide sufficient evidence, without requiring additional exploration; second, RJE retrieves key paths from the KG and begins exploration from strategically selected entities along these paths, rather than starting from topic entities as in ToG and PoG. This early resolution capability and targeted exploration approach significantly reduce computational demands while maintaining high accuracy. 
Experimental results in Appendix~\ref{sec:Analysis of Exploration Round Reduction} demonstrate that initiating exploration from selected entities along retrieval paths reduces exploration rounds, thereby alleviating computational burden.

\section{Conclusion}

In this paper, we introduce the RJE framework, which addresses core limitations in existing KGQA methods by integrating precise retrieval, sufficiency judgment, and conditional exploration. Our specialized auxiliary modules, including Reasoning Path Ranking, Question Decomposition, and Retriever-assisted Exploration, significantly enhance the reasoning efficacy of smaller open-source LLMs in knowledge-intensive tasks. Empirical evaluations across standard KGQA benchmarks demonstrate that RJE not only outperforms existing approaches in both accuracy and efficiency, but also enables small-sized open-source LLMs to achieve comparable performance, advancing the development of more efficient and accessible KGQA systems.

\section*{Limitations}
While our approach demonstrates significant improvements in KGQA, there are several limitations that suggest directions for future work. First, existing KGs, which are primarily constructed from internet corpora, often contain noisy triples and outdated information. Such noisy knowledge can mislead LLMs into making incorrect responses, even when using our proposed framework. In future work, we plan to investigate methods for detecting and filtering unreliable knowledge in KGs to reduce noise and enhance KGQA system reliability. Second, our experimental evaluation is limited to English language datasets. To assess the cross-lingual capabilities of our approach, we intend to extend our evaluation to multiple languages. 

\section*{Acknowledgments}
The research was partially supported by ``Pioneer'' and ``Leading Goose'' R\&D Program of Zhejiang 2023C01029, the Plans for Major Provincial Science\&Technology Projects No.:202303a07020006, and the China National Natural Science Foundation with no. 62132018.

\bibliography{custom}

\appendix

\section{Datasets}
\label{sec:Datasets}

We evaluate the proposed method on the WebQuestionsSP and ComplexWebQuestions datasets. For fair comparison with prior research \citep{sunthink,chenplan}, we maintain identical training and testing splits. Table~\ref{tab:kgqa_datasets} presents a comprehensive summary of the dataset statistics.

\section{Baselines}
\label{sec:appendix_Baselines}

(1) \textbf{LLM-only methods}, these approaches rely solely on large language models without explicit knowledge graph integration:

\textit{Standard prompting (IO prompt)} \citep{brown2020language} demonstrated that LLMs outperform traditional language models on task-agnostic and few-shot problems.

\textit{Chain-of-Thought prompting (CoT)} \citep{wei2022chain} incorporates ``think step by step'' prompts to enhance LLM performance across various natural language processing tasks.

\textit{Self-Consistency (SC)} \citep{wangself} improves performance by sampling multiple diverse reasoning paths through few-shot CoT prompts and selecting the most consistent answer among the generated outputs.

\begin{table}[t]
\centering
\resizebox{\columnwidth}{!}
{
\begin{tabular}{llcccc}
\toprule
\textbf{Dataset} & \textbf{KG} & \textbf{Train} & \textbf{Dev} & \textbf{Test} & \textbf{Max hop}\\
\midrule
WebQSP   & Freebase & 2,848  & 250    & 1,639 & 2\\
CWQ      & Freebase & 27,639 & 3,519  & 3,531 & 4\\

\bottomrule
\end{tabular}
}
\caption{Statistics of different KGQA datasets used in the experiments.}
\label{tab:kgqa_datasets}
\end{table}

(2) \textbf{Retrieval-Reasoning Methods}, these methods focus on effective subgraph retrieval techniques:

\textit{SR} \citep{zhang2022subgraph} proposes a trainable subgraph retriever decoupled from the downstream reasoner, incorporating a PLM to expand paths for subgraph induction with automatic termination criteria.

\textit{UniKGQA} \citep{jiangunikgqa} integrates graph retrieval and reasoning into a single model that incorporates a PLM.

\textit{ReasoningLM} \citep{jiang2023reasoninglm} enables effective question understanding and structured reasoning over knowledge graphs by incorporating subgraph-aware self-attention and an adaptation tuning strategy.

\textit{EPR} \citep{ding2024enhancing} enhances subgraph extraction for KGQA by modeling and retrieving structural evidence patterns that connect necessary entities and relations. 

(3) \textbf{Retrieval-Augmented Generation Methods}, these approaches combine retrieval mechanisms with LLM capabilities:

\textit{KG-CoT} \citep{zhao2024kg} enhances LLMs reasoning by integrating step-by-step graph reasoning over KGs to generate explicit reasoning paths, enabling knowledge-aware, plug-and-play CoT prompting.

\textit{RD-P} \citep{huang2024rd} integrates KGs with LLMs by retrieving and verifying trustworthy reasoning paths to construct reliable prompts, enhancing reasoning accuracy and efficiency without modifying LLM parameters.

\textit{SubgraphRAG} \citep{li2025simple} enhances KG-based RAG by efficiently retrieving high-quality subgraphs using a lightweight MLP with structural features, enabling effective and adaptable LLM reasoning without fine-tuning.

(4) \textbf{Fine-tuned Agent-based Methods}, these methods involve fine-tuning LLMs with knowledge graph information:

\textit{RoG} \citep{luoreasoning} enables faithful and interpretable KG reasoning by fine-tuning LLMs within a planning-retrieval-reasoning framework that distills knowledge into relation path planning and optimized reasoning over retrieved KG paths.

\textit{KG-Agent} \citep{jiang2024kg} empowers small-sized LLMs to autonomously perform complex KG reasoning by fine-tuning them with code-based instruction data and integrating a multifunctional toolbox for structured operations and iterative decision-making.

\textit{GCR} \citep{luo2024graph} ensures faithful KG reasoning by fine-tuning a lightweight KG-specialized LLM to generate constrained decoding paths via a KG-Trie index, then leveraging a general inductive power of LLM to produce accurate final answers.

(5) \textbf{Prompting Agent-based Methods}, these approaches utilize sophisticated prompting techniques with KG integration:

\textit{StructGPT} \citep{jiang2023structgpt} defines an interface for accessing and filtering knowledge from KGs data under limited constraints, and leverages LLMs to iteratively infer answers or generate follow-up plans.

\textit{Interactive-KBQA} \citep{xiong2024interactive} Interactive KBQA directly leverages LLMs to interact with KGs, subsequently generating logical forms.

\textit{ReKnoS} \citep{wang2025reasoning} enhances knowledge graph reasoning by introducing super-relations, which group domain-specific relations to expand the reasoning space and improve retrieval performance.

\textit{ToG} \citep{sunthink} iteratively retrieves relevant triples from the knowledge graph and uses LLMs to evaluate whether the reasoning paths within beam search are sufficient to answer the question or if additional next-hop information is needed.

\textit{PoG} \citep{chenplan} builds upon ToG by incorporating mechanisms such as Adaptive Breadth, Guidance, Memory, and Reflection.

\section{Analysis of Macro-F1 Metric}
\label{sec:Analysis of Macro-F1 Metric}

\begin{table}[t]
  \centering
  % \small
  \resizebox{\columnwidth}{!}{
  \begin{tabular}{lcccc}
    \hline
    \multirow{2}{*}{\textbf{Method}} & \multicolumn{2}{c}{\textbf{CWQ}} & \multicolumn{2}{c}{\textbf{WebQSP}} \\
    \cline{2-3}\cline{4-5}
     & \textbf{Hit@1} & \textbf{Macro-F1} & \textbf{Hit@1} & \textbf{Macro-F1} \\
    \hline

    \multicolumn{5}{c}{\textit{Baselines}} \\
    \hline
    SR \citep{zhang2022subgraph}           & 50.2 & 47.1 & 68.9 & 64.1 \\
    UniKGQA \citep{jiangunikgqa}           & 51.2 & 49.0 & 77.2 & 72.2 \\
    ReasoningLM \citep{jiang2023reasoninglm}& 69.0 & 64.0 & 78.5 & 71.0 \\
    EPR \citep{ding2024enhancing}          & 60.6 & 61.2 & 71.2 & 70.2 \\
    RD-P w/ChatGPT \citep{huang2024rd}     & 63.5 & 56.6 & 85.3 & 69.7 \\
    SubgraphRAG w/GPT4o \citep{li2025simple}& 67.5 & 59.5 & 90.9 & 78.2 \\
    RoG \citep{luoreasoning}               & 62.6 & 56.2 & 85.7 & 70.8 \\
    KG-Agent \citep{jiang2024kg}           & 72.2 & 69.8 & 83.3 & \textbf{81.0} \\
    GCR \citep{luo2024graph}               & 75.8 & 61.7 & 92.2 & 74.1 \\
    \hline

    \multicolumn{5}{c}{\textit{Ours}} \\
    \hline
    RJE w/Llama3.2-3B   & 62.9 & 53.7 & 82.6 & 65.9 \\
    RJE w/Llama3.1-8B   & 71.5 & 60.5 & 89.2 & 73.8 \\
    RJE w/Qwen2.5-14B   & 71.2 & 62.8 & 90.3 & 76.0 \\
    RJE w/ChatGPT       & 72.6 & 62.1 & 91.2 & 74.8 \\
    RJE w/GPT-4o-mini   & 77.1 & 65.9 & \textbf{92.5} & 77.6 \\
    RJE w/DeepSeek-V3   & \textbf{78.2} & \textbf{70.2} & 92.4 & 78.9 \\
    \hline
  \end{tabular}
  }
  \caption{Results on CWQ and WebQSP under Hits@1 and Macro-F1; the best results are highlighted in bold.}
  \label{tab:f1_score}
\end{table}

To provide a more comprehensive assessment of RJE's performance, we conduct additional evaluations using Macro-F1 across various LLMs. Since Macro F1 is not consistently reported in prior work (e.g., PoG~\citep{chenplan} and ToG~\citep{sunthink}), we compare against only those baselines that include this metric.

As shown in Table~\ref{tab:f1_score} , RJE achieves superior performance compared to most baselines on both Hits@1 and Macro-F1.  The only exception occurs on the WebQSP dataset, where KG-agent (a fine-tuned LLM-based method) marginally outperforms RJE with DeepSeek-V3 on Macro-F1. However, RJE with DeepSeek-V3 significantly surpasses KG-agent in Hits@1 performance. The excellent Macro-F1 performance of RJE indicates that it simultaneously achieves high precision and recall in knowledge-graph reasoning tasks.

\section{The Performance of Reasoning Path Ranking}
\label{sec:The Performance of Reasoning Path Ranker}

\begin{figure}[htbp]
  \centering
  \begin{subfigure}[b]{0.9\linewidth}  % 让子图宽一点，看着舒服
      \centering
      \includegraphics[width=\linewidth]{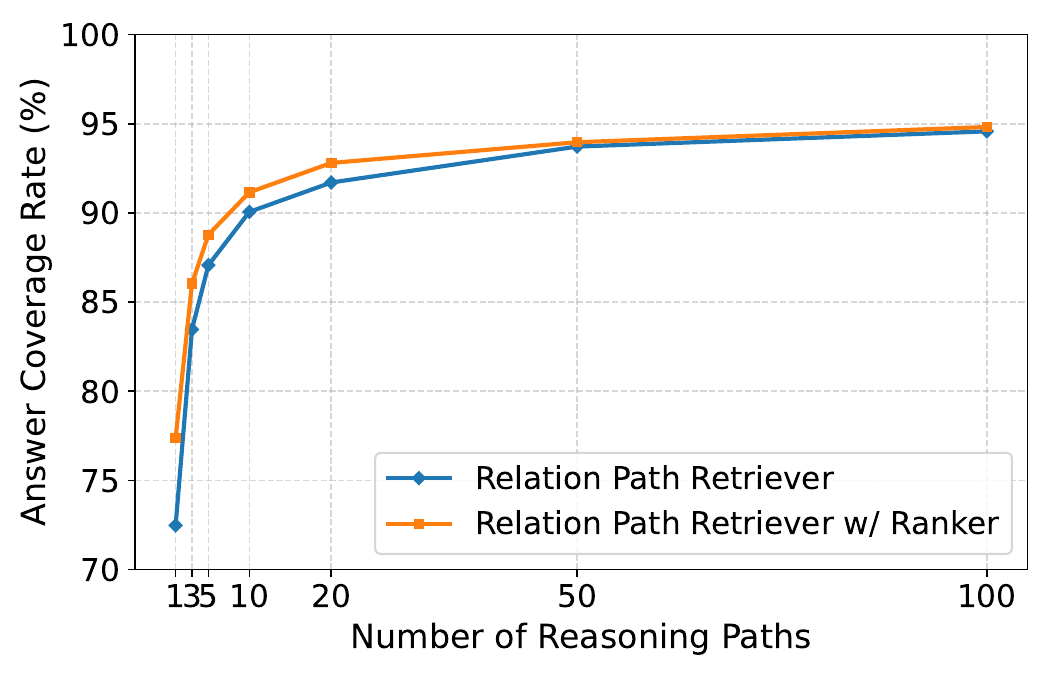}
      \caption{WebQSP}
  \end{subfigure}
  
  \vspace{0.5em} % 两张图之间加一点垂直间距
  
  \begin{subfigure}[b]{0.9\linewidth}
      \centering
      \includegraphics[width=\linewidth]{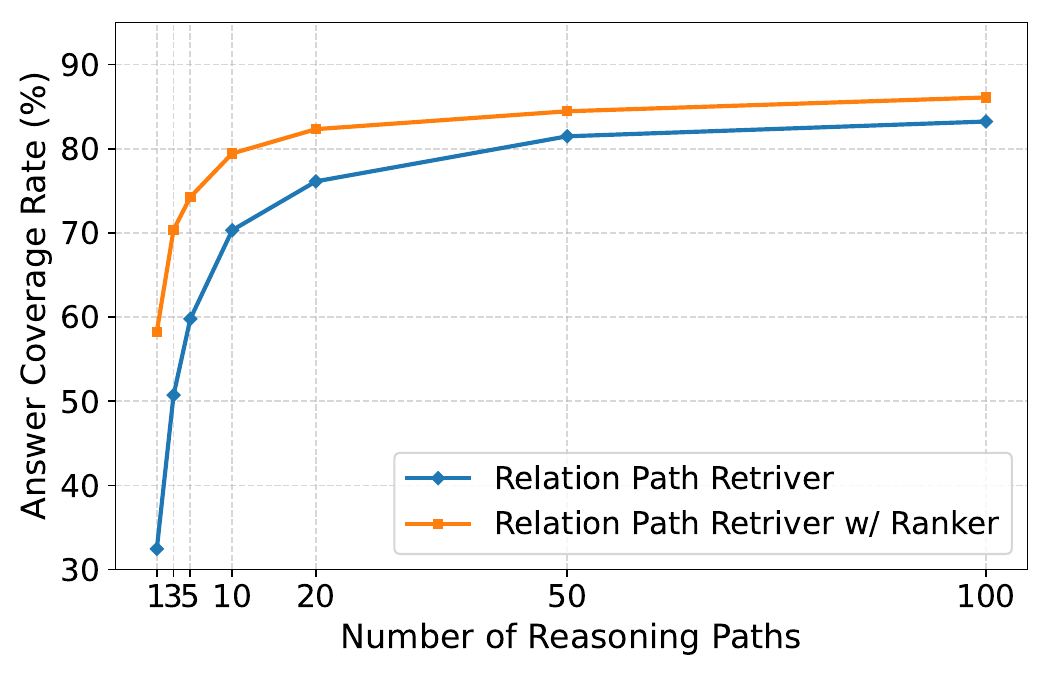}
      \caption{CWQ}
  \end{subfigure}
  
  \caption{Comparison of Answer Coverage Across Different Numbers of Candidate Reasoning Paths.}
  \label{fig:Ranker_Performance}
\end{figure}

We conduct a systematic evaluation of the impact of the reasoning path ranking on retrieval performance, using answer coverage at various retrieval scales as the metric. Figure~\ref{fig:Ranker_Performance} presents a comparative analysis between the standalone relation retriever and its integration with the reasoning path ranker across two datasets.
The results demonstrate that our combined approach consistently achieves higher coverage across all candidate reasoning path sizes. Notably, on the more challenging CWQ dataset, the contribution of the ranker yields particularly substantial improvements. Furthermore, as the candidate path count decreases, the performance enhancement provided by the ranker becomes increasingly significant.

\section{Addition Ablation Study}
\label{sec:Addition Ablation study}

The RJE framework contains three manually configurable hyperparameters: the number of reasoning paths to be refined by the reasoning path ranker, the number of relations to be considered by the relation path retriever during retriever-assisted exploration and the maximum number of exploration rounds. To investigate how these parameters influence the performance of RJE, we conduct a series of experiments and derive practical insights.

\begin{figure}[t]
  \includegraphics[width=\columnwidth]{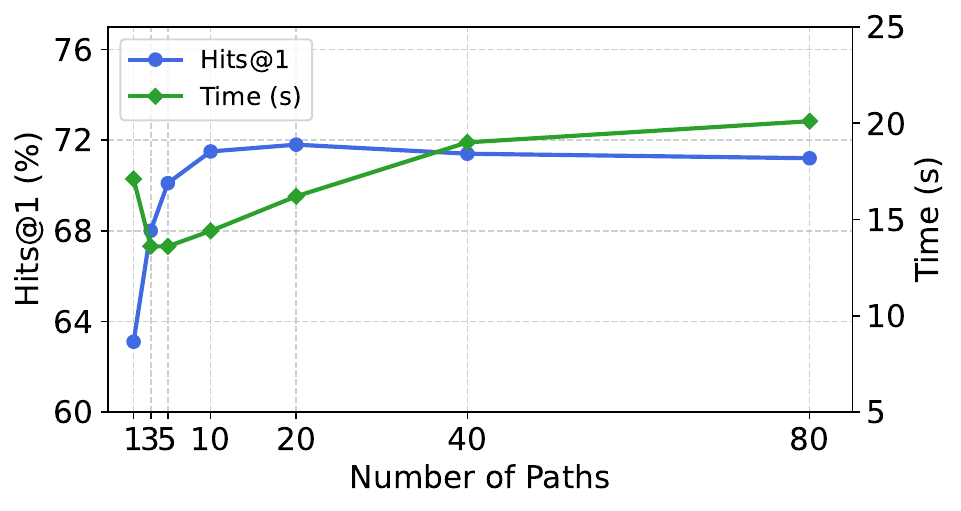}
  \caption{The impact of the number of Reasoning Paths on performance on the CWQ dataset.}
  \label{fig:path_numm}
\end{figure}

\begin{figure}[t]
  \includegraphics[width=\columnwidth]{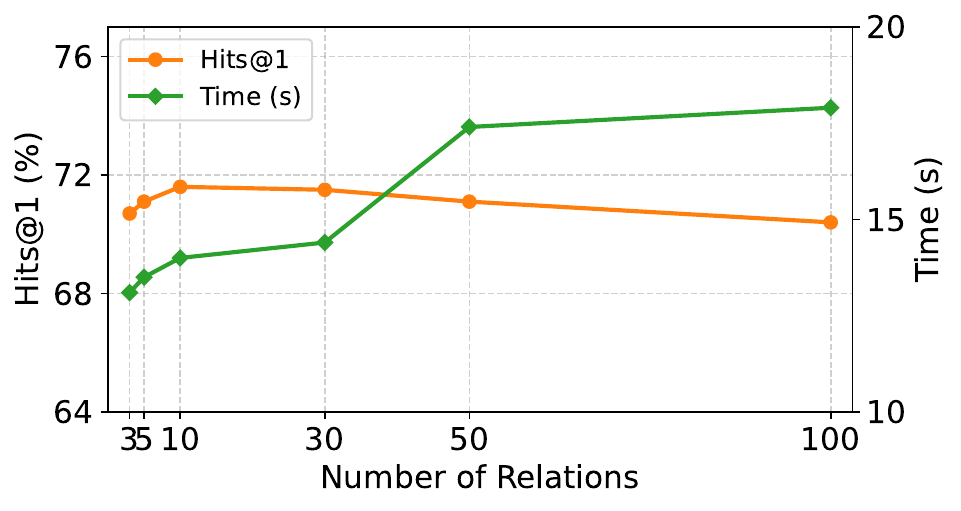}
  \caption{The impact of the number of Relations on performance on the CWQ dataset.}
  \label{fig:rel_num}
\end{figure}

\begin{figure}[t]
  \includegraphics[width=\columnwidth]{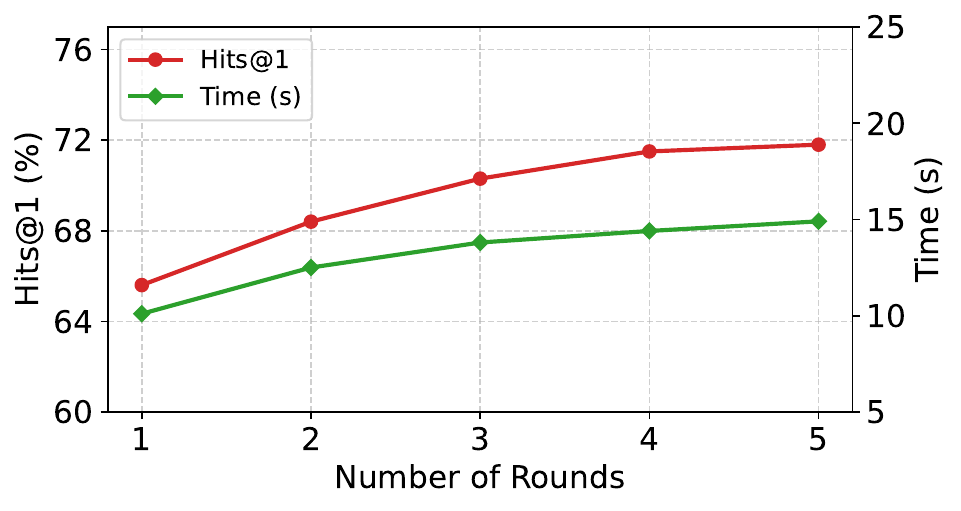}
  \caption{The impact of exploration rounds on performance on the CWQ dataset.}
  \label{fig:dep_num}
\end{figure}

\subsection{Impact of the Number of Paths}

In the RJE framework, the reasoning path ranker outputs top-$K$ candidate paths for judgment evaluation. The number of paths represents a critical trade-off: too few paths limit available evidence, while too many introduce noise that degrades LLM performance. To examine this balance, we conduct experiments on CWQ with varying path counts.

As shown in Figure~\ref{fig:path_numm}, performance increases substantially from 1 to 10 paths, confirming the ranker effectively identifies relevant information for accurate judgment. However, performance declines beyond 20 paths, indicating that excessive candidates introduce irrelevant information that misleads the LLM. The relationship between path count and time cost exhibits a U-shaped pattern. Initially, increasing path count reduces overall time as sufficient evidence from multiple paths decreases the need for extensive reasoning exploration. However, when the path count becomes too large, the additional tokens and noise both mislead the LLM and increase processing time.

These results demonstrate that the choice of path count significantly affects performance. An appropriate setting balances the provision of sufficient evidence with noise reduction, improving both reasoning accuracy and computational efficiency.

\subsection{Impact of the Number of Relations}

Previous agent-based approaches typically feed all relations into LLMs during relation exploration. Although effective with powerful proprietary LLMs, this strategy often degrades performance in smaller open-source models due to excessive input length. To address this limitation, we employ the relation path retriever to pre-filter the $N$ most relevant relations, which are then fed to the LLMs for final selection. Notably, $N$ is smaller than the total number of relations. We conduct experiments on CWQ with varying $N$ values to assess the performance impact. 

As shown in Figure~\ref{fig:rel_num}, varying the relation count does not cause large performance fluctuations because the retriever consistently ranks relevant relations at the top. Even a small number of relations can yield strong performance. On closer inspection, increasing the number of relations from 3 to 10 yields gradual gains. However, beyond $N=10$, performance begins to decline, suggesting that excessive relations introduce noise that impairs the LLM’s focus on the most informative candidates. Processing time increases with relation count due to additional tokens and LLM exploration.

These results demonstrate that the relation path retriever effectively narrows the exploration space while enhancing LLM reasoning capabilities through focused candidate provision.

\subsection{Impact of Exploration Rounds}

To prevent LLMs from engaging in endless exploration, we set an upper limit on the number of exploration rounds. We conduct experiments on CWQ with varying numbers of rounds. As shown in Figure~\ref{fig:dep_num}, performance consistently improves with additional rounds, suggesting that further exploration enables RJE to discover more relevant information for answering complex questions. However, after four rounds, the performance gain becomes marginal, since the maximum hop count on CWQ is 4.

Processing time increases with the maximum round due to additional exploration cycles. However, the time overhead exhibits diminishing gains at higher values, as fewer questions actually require extensive exploration rounds. To balance computational cost and effectiveness, we set the maximum exploration rounds to 4.

\section{Results Analysis}
\label{sec:Results Analysis}

\begin{figure}[htbp]
  \centering
  \begin{subfigure}[b]{0.9\linewidth}  % 让子图宽一点，看着舒服
      \centering
      \includegraphics[width=\linewidth]{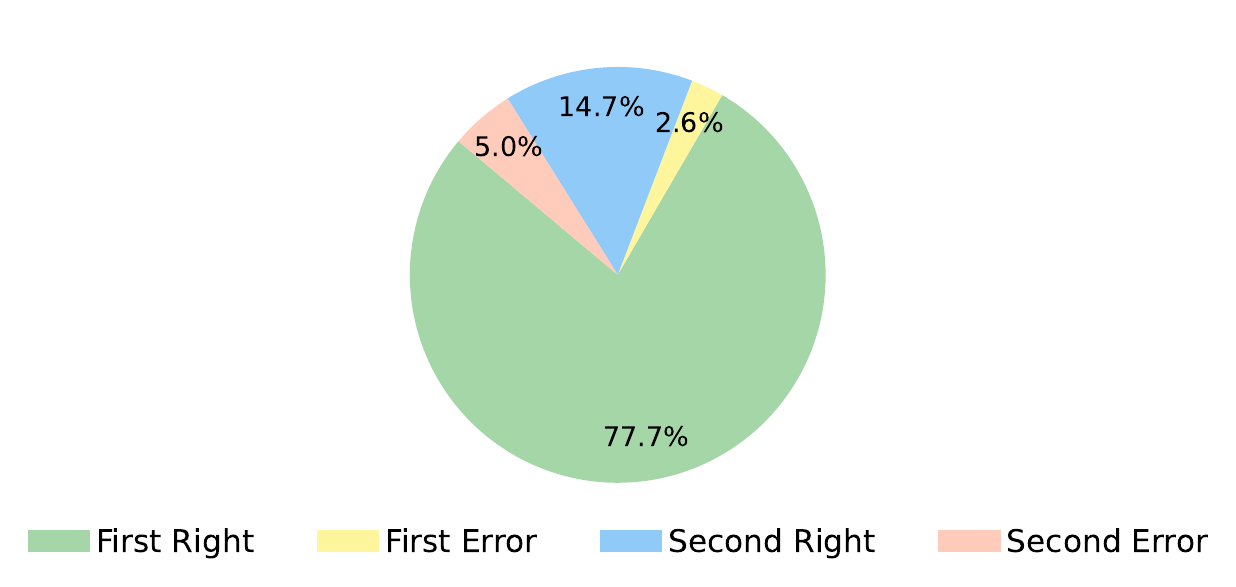}
      \caption{WebQSP}
  \end{subfigure}
  
  % \vspace{0.5em} % 两张图之间加一点垂直间距
  
  \begin{subfigure}[b]{0.9\linewidth}
      \centering
      \includegraphics[width=\linewidth]{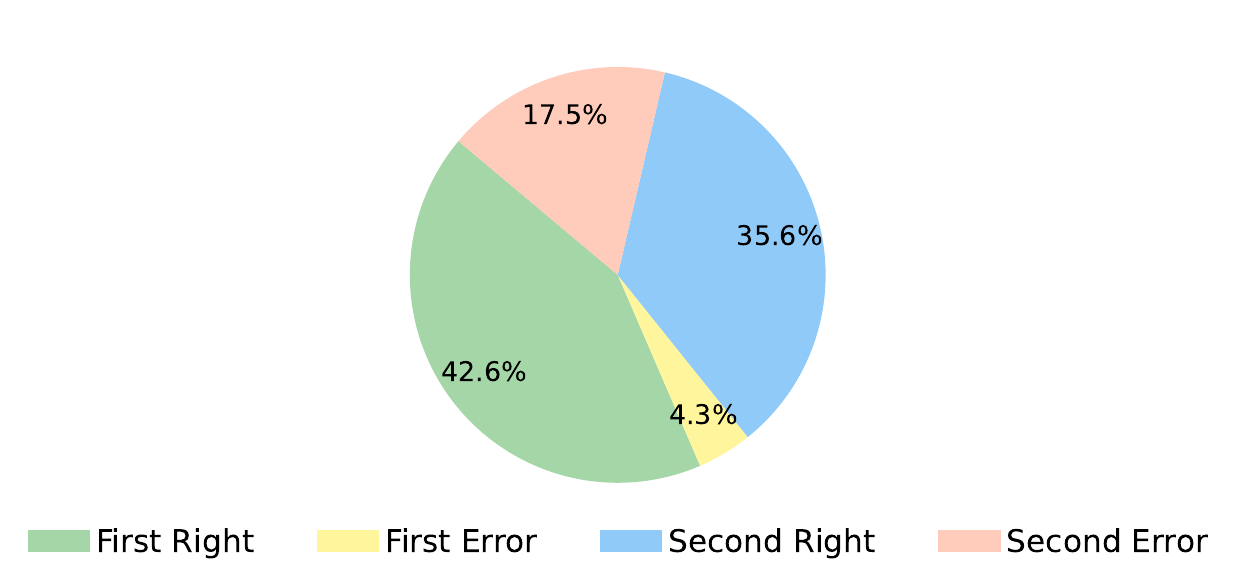}
      \caption{CWQ}
  \end{subfigure}
  
  \caption{Answer accuracy at different stages in the RJE framework on WebQSP and CWQ.}
  \label{fig:Deepseek_result}
\end{figure}

In the RJE framework, the final answer can be produced either during the judgment stage or the exploration stage. Figure~\ref{fig:Deepseek_result} illustrates the answer accuracy at each stage using DeepSeek-V3 on two datasets. The label \textbf{First} denotes answers generated during the judgment stage, while \textbf{Second} indicates those generated during the exploration stage.

For WebQSP, 77.7\% of the questions are correctly answered during the judgment stage, indicating that only a single LLM call is required. This demonstrates the efficiency of RJE in producing accurate answers with minimal computational cost when the path information is sufficient. Among the remaining 19.7\% of cases where the initial paths are insufficient, 14.7\% of the questions are still answered correctly during the subsequent exploration stage. These findings demonstrate the capacity of RJE to recognize insufficient information and retrieve additional knowledge through exploration, eventually leading to the correct answer.

For CWQ, which is generally more complex than WebQSP, approximately half of the questions are answered during the judgment stage. This outcome reflects the ability of RJE to adapt its strategy according to the complexity of the question. Notably, 35.6\% of the questions are correctly answered during the exploration stage, demonstrating the potential of RJE in handling more challenging questions by exploring beyond the initially retrieved paths.

\section{Case Study}
\label{sec:Case Study}

\subsection{Case 1}

\begin{figure*}[htbp]
  \centering
  \includegraphics[width=\textwidth]{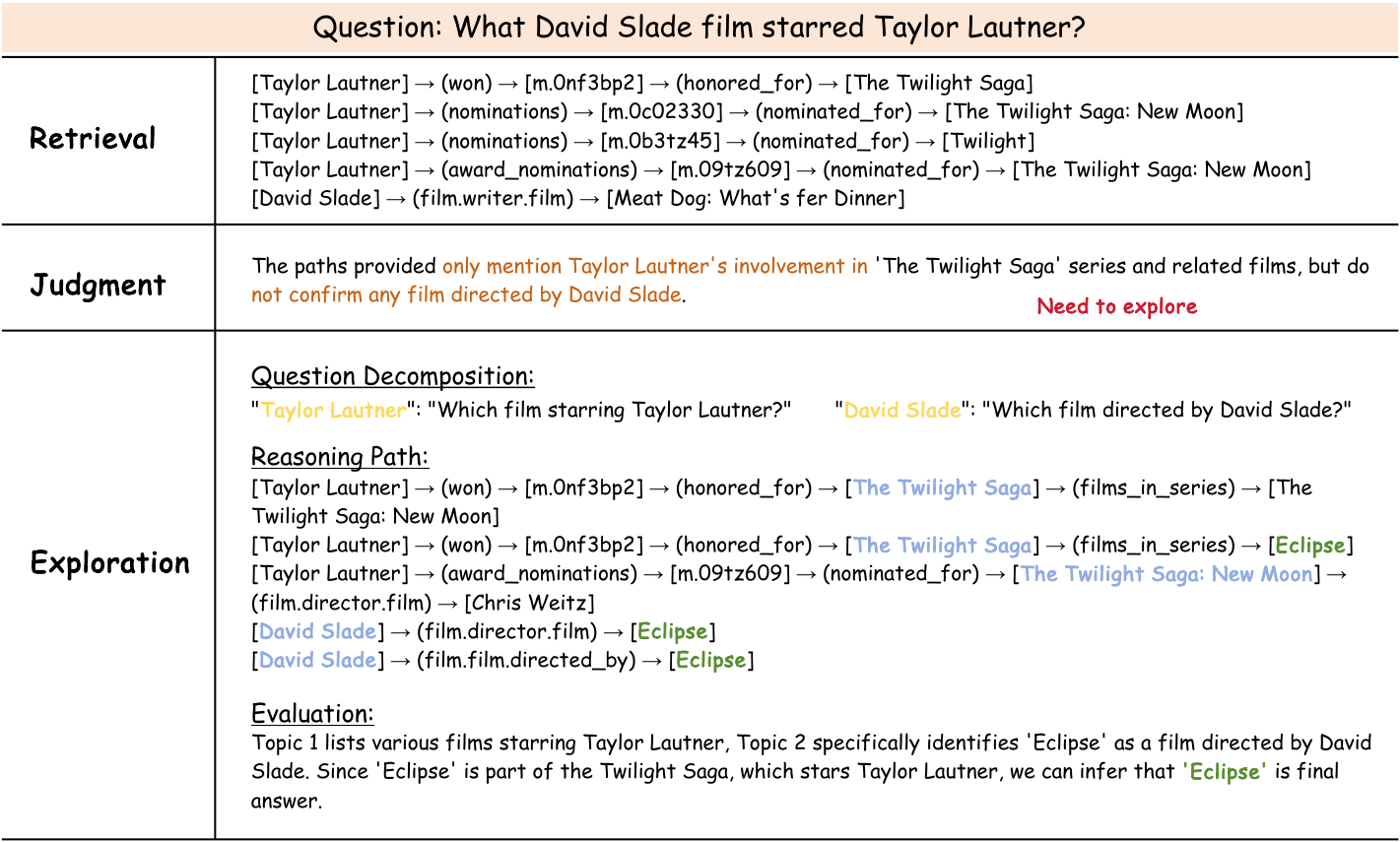}
  \caption{A typical case demonstrates the advantages of the RJE framework. The topic entities, exploration entities, incorrect answers, and correct answers are highlighted in yellow, blue, red, and green, respectively.}
  \label{fig:case_1}
\end{figure*}

Figure~\ref{fig:case_1} illustrates a case from CWQ that highlights the advantages of the RJE framework. For the question \textit{``What David Slade film starred Taylor Lautner?''}, RJE first retrieves and refines five highly relevant paths. After performing judgment on these paths, RJE identifies that while some of the paths indicate movies involving \textit{``Taylor Lautner''} and one indicates a movie directed by \textit{``David Slade''}, there is no overlapping entity between them that can support valid reasoning. Therefore, RJE judges that the path information is insufficient and triggers further exploration.

\begin{figure*}[htbp]
  \centering
  \includegraphics[width=\textwidth]{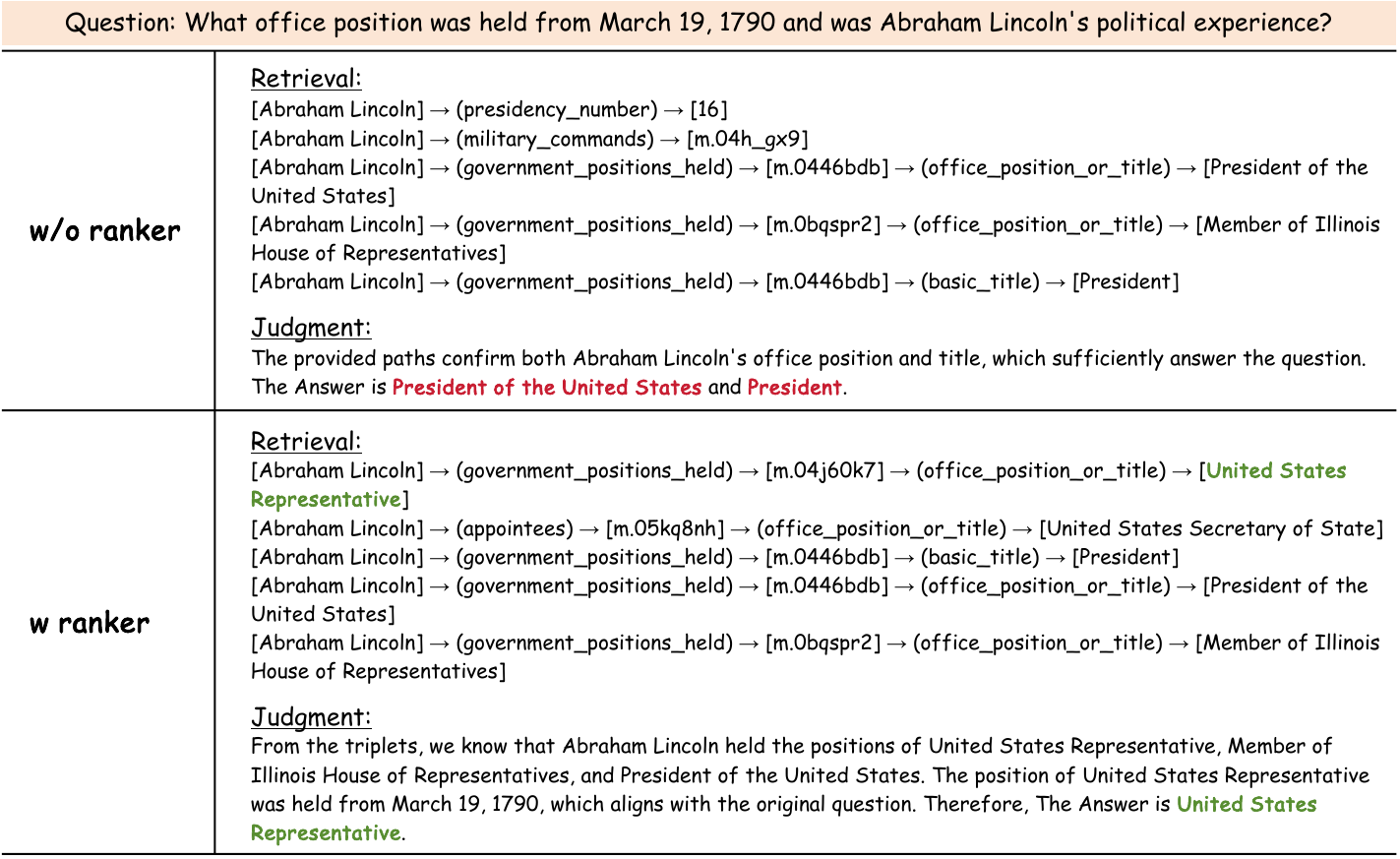}
  \caption{A typical case illustrates the importance of the reasoning path ranker. With and without the ranker, the LLM infers different answers. The incorrect and correct answers are highlighted in red and green, respectively.}
  \label{fig:case_2}
\end{figure*}

Upon entering the exploration stage, RJE first performs question decomposition based on the topic entities to better facilitate path exploration and reasoning. In the first round of exploration, RJE successfully selects the entities \textit{``The Twilight Saga''}, \textit{``The Twilight Saga: New Moon''}, and \textit{``David Slade''}, aiming to retrieve detailed information about the movie series and its director. After conducting relation and entity exploration, RJE acquires key knowledge: \textit{``Eclipse''} is part of \textit{``The Twilight Saga''}, and \textit{``David Slade''} is its director. By synthesizing this information, RJE is able to accurately infer that the answer to the original question is \textit{``Eclipse''}.

In summary, while three hops would be needed to reach the correct answer theoretically, RJE obtains highly relevant information during the retrieval stage, eliminating the need to start exploration from the topic entity and significantly reducing the number of exploration rounds. This substantially reduces computational cost and inference time. Furthermore, the integration of question decomposition and the retriever-assisted exploration during the exploration process enables RJE to perform path exploration and reasoning more effectively, making it easier to arrive at the correct answer. These strengths contribute to the strong performance of RJE even when using LLMs with relatively small parameter sizes.

\subsection{Case 2}
Figure~\ref{fig:case_2} presents another case from CWQ that demonstrates the importance of the reasoning path ranker. For the question \textit{``What office position was held from March 19, 1790 and was Abraham Lincoln's political experience?''}, without the reasoning path ranker, RJE retrieves the top five paths with the highest relevance scores. However, none of these paths cover the correct answer. As a result, when the LLM analyzes these paths, it is misled by the weakly relevant and noisy information, leading to an incorrect answer during the judgment stage.

In contrast, with the reasoning path ranker reordering the paths obtained by the relation path retriever, the paths containing the correct answer are successfully selected and ranked at the top. Upon analyzing the reordered paths, the LLM observes the entity \textit{``United States Representative''}, and, based on its internal knowledge that the United States Representative office was held from March 19, 1790, it aligns this information with the question and infers the correct answer.

In summary, the reasoning path ranker effectively filters and prioritizes the most relevant paths, helping LLMs reduce hallucinations and enhancing the probability of answering correctly during the judgment stage without needing further exploration. This substantially reduces computational cost and inference time. Overall, RJE dynamically adjusts its strategy based on question complexity and path relevance, aiming to find the correct answer while reducing resource consumption.

\section{Analysis of Exploration Round Reduction}
\label{sec:Analysis of Exploration Round Reduction}
To validate that RJE improves exploration efficiency by starting from entities identified in retrieval paths, we conducted experiments on a subset of CWQ questions that explicitly require exploration. This subset excludes questions that RJE can directly answer during the judgment stage, ensuring all questions necessitate exploration. We compared methods using their default maximum number of rounds: RJE (4 rounds), PoG (4 rounds) and ToG (3 rounds). RJE w/o Retrieval is a variant without retrieval paths that initiates exploration from topic entities rather than retrieval path entities. 

As shown in Table~\ref{tab:step_reduction}, RJE achieves the lowest average number of rounds (1.57), substantially outperforming RJE w/o Retrieval (2.41 rounds) by 35\%. This indicates that starting exploration from retrieval path entities effectively reduces exploration rounds. RJE also outperforms PoG (2.36 rounds) and ToG (2.58 rounds), highlighting the computational efficiency of the RJE framework.

\begin{table}[t]
\centering
\resizebox{\columnwidth}{!}
{
\begin{tabular}{lcccc}
\toprule
\textbf{Method} & \textbf{RJE} & \textbf{RJE w/o Retrieval} & \textbf{PoG} & \textbf{ToG} \\
\midrule
\textbf{Average Rounds} & \textbf{1.57} & 2.41 & 2.36 & 2.58 \\
\bottomrule
\end{tabular}
}
\caption{Average number of exploration rounds required by different methods on CWQ subset.}
\label{tab:step_reduction}
\end{table}

\section{Performance on Different Hops}

To evaluate RJE's performance across questions of varying complexity, we analyze results by hop count on the CWQ dataset. 
For each question, the hop count is defined as the longest of the shortest paths from topic entities to the answer entity in the provided SPARQL queries.

As shown in Table~\ref{tab:performance_hops}, RJE maintains consistent performance across different hop requirements, achieving Hits@1 scores of 69.9\%-81.5\%. This consistency highlights RJE’s effectiveness on multi-hop reasoning, attributed to its systematic three-stage architecture that mitigates the difficulty of multi-hop questions.

\begin{table}[t]
\centering
% \resizebox{\columnwidth}{!}
% {
\begin{tabular}{lcccc}
\toprule
\textbf{CWQ} & \textbf{1 hop} & \textbf{2 hop} & \textbf{3 hop} & \textbf{4 hop} \\
\midrule
\textbf{Hits@1} & 73.7 & 80.0 & 81.5 & 69.9 \\
\bottomrule
\end{tabular}
% }
\caption{RJE performance on questions with different numbers of hops using DeepSeek-V3.}
\label{tab:performance_hops}
\end{table}

% \onecolumn
\section{Prompts}
\subsection{Judgment} \label{prompt_Paths Evaluation}

\begin{tcolorbox}[breakable, enhanced]

Your task is to infer the answer based on the given question and given triple paths.

\{In-Context Few-shot\}

Task Requirements:

1. Do not provide explanations or extra text.

2. The output must be in strict JSON format.

Now, based on the following input, determine whether the question can be answered.

Question:

Paths:

\end{tcolorbox}

\subsection{Question Decomposition} \label{prompt_Question Decomposition}
\begin{tcolorbox}[breakable, enhanced]

You are an expert in question decomposition and knowledge-based reasoning.

Given:

- A complex natural language question.

- A list of topic entities mentioned in the question.

Your tasks are: Generate a sub-question for each entity. The sub-question should reflect the original question's intent, but be scoped only to that specific entity.

\{In-Context Few-shot\}

Original Question:

Topic Entities:
\end{tcolorbox}

\subsection{Exploration Entities Selection} \label{prompt_Exploration Entities Selection}
\begin{tcolorbox}[breakable, enhanced]
Your task is to determine which entities should be explored next, based on the Original Question, Topic Question and given triple paths.

\{In-Context Few-shot\}

Now you need to output the entities from the Entity List, without additional explanations or formatting. Strictly follow the entity names as they appear in the Entity List.

Note: Only include entities that are necessary for answering the Original Question, and ensure the list has no more than 10 entities.

Original Question:

Topic 1:

Topic Question:

Topic Entity:

Triplets:

Topic 2:

…
\end{tcolorbox}

\subsection{Relation Exploration}  \label{prompt_Relation Exploration}
\begin{tcolorbox}[breakable, enhanced]
Your task is to select useful relations from a given list based on the current question and connected entity.

\{In-Context Few-shot\}

Task Requirements:

1. Strictly output only the selected relations from the provided list.

2. Do not include any additional relations, explanations, reasoning, or extra formatting.

Now, based on the following input, select the useful relations.

Question:

Connected entity:

Relations List:

\end{tcolorbox}

\subsection{Entity Exploration} \label{prompt_Entity Exploration}
\begin{tcolorbox}[breakable, enhanced]
Your task is to select the minimal set of relevant entities from the given triplets.

\{In-Context Few-shot\}

Task Requirements:

1. Entities must come strictly from the given list; do not introduce new entities.

2. Strictly output only the selected entities, without explanations or additional formatting.

Now, based on the following input, select the minimal relevant entities.

Question:

Triplets:

\end{tcolorbox}

\subsection{Answer Generation} \label{prompt_Answer Generation}
\begin{tcolorbox}[breakable, enhanced]
Your task is to infer the answer to the original question by first reasoning over each Topic Question using the provided triplets and your knowledge, then combining the insights from all Topic Questions to derive the final answer.

Instructions:

1. For each topic entity:
   
   - Read its corresponding topic question.
   
   - Use the associated triplets and your knowledge to infer an answer.

2. After processing all topic entities:
  
   - Analyze how the individual answers relate to the original question.
   
   - If possible, synthesize them to derive the final answer.

\{In-Context Few-shot\}

Task Requirements:

1. Do not provide explanations or extra text.

2. The output must be in strict JSON format.

Now, based on the following input, determine whether the question can be answered.

Original Question:

Topic 1:

Topic Question:

Topic Entity:

Triplets:

Topic 2:

…

\end{tcolorbox}

\section{SPARQL}
\subsection{Relation Search} \label{Relation_SPARQL}
\begin{tcolorbox}[breakable, enhanced,colback=white]
PREFIX ns: <http://rdf.freebase.com/ns/>

SELECT DISTINCT ?relation

WHERE \{  

\hspace*{2em}ns:\%s ?relation ?x . 

\hspace*{2em}FILTER (?x != ns:\%s)

\}
\end{tcolorbox}

\begin{tcolorbox}[breakable, enhanced,colback=white]
PREFIX ns: <http://rdf.freebase.com/ns/>

SELECT DISTINCT ?relation

WHERE \{  

\hspace*{2em}?x ?relation ns:\%s . 

\hspace*{2em}FILTER (?x != ns:\%s)

\}
\end{tcolorbox}

\subsection{Entity Search} \label{Entity_SPARQL}
\begin{tcolorbox}[breakable, enhanced, colback=white]
PREFIX ns: <http://rdf.freebase.com/ns/>

SELECT ?tailEntity

WHERE \{

\hspace*{2em}ns:\%s ns:\%s ?tailEntity .

\}
\end{tcolorbox}

\begin{tcolorbox}[breakable, enhanced, colback=white]
PREFIX ns: <http://rdf.freebase.com/ns/>

SELECT ?tailEntity

WHERE \{

\hspace*{2em}?tailEntity ns:\%s ns:\%s .

\}
\end{tcolorbox}

\end{document}